%% file: main.tex
\documentclass[10pt,twocolumn,letterpaper]{article}

\usepackage{cvpr}              % To produce the CAMERA-READY version

\definecolor{cvprblue}{rgb}{0.21,0.49,0.74}
\usepackage[pagebackref,breaklinks,colorlinks,allcolors=cvprblue]{hyperref}
\usepackage{hyperref}
\usepackage{url}
\usepackage{multirow,booktabs,graphicx,siunitx,wrapfig,subcaption}
\usepackage{colortbl}
\usepackage{xcolor}
\usepackage{multicol}
\usepackage{balance}

\def\paperID{***} % *** Enter the Paper ID here
\def\confName{CVPR}
\def\confYear{2025}

\title{Text-guided Sparse Voxel Pruning for Efficient 3D Visual Grounding}

\author{Wenxuan Guo\textsuperscript{1$*$} \quad 
Xiuwei Xu\textsuperscript{1$*$} \quad
Ziwei Wang\textsuperscript{2} \quad
Jianjiang Feng\textsuperscript{1$\dagger$} \quad
Jie Zhou\textsuperscript{1} \quad
Jiwen Lu\textsuperscript{1$\dagger$} \\
	\textsuperscript{1}Tsinghua University \qquad \textsuperscript{2}Nanyang Technological University\\
	{\tt\small\{gwx22,xxw21\}@mails.tsinghua.edu.cn \quad ziwei.wang@ntu.edu.sg} \\
	{\tt\small \{jfeng,jzhou,lujiwen\}@tsinghua.edu.cn}}
\begin{document}
\maketitle
{
        \renewcommand{\thefootnote}{*}
	\footnotetext[5]{Equal contribution. $^{\dagger}$ Corresponding author.}
}

\input{sec/0_abstract}    
\input{sec/1_intro}
\input{sec/2_related}
\input{sec/3_method}
\input{sec/4_exper}
\input{sec/5_conclu}

% \clearpage
% \clearpage

% \balance
{
    \small
    \bibliographystyle{ieeenat_fullname}
    \bibliography{main}
}
\input{sec/X_suppl}


\end{document}

%% file: sec/0_abstract.tex
\begin{abstract}
In this paper, we propose an efficient multi-level convolution architecture for 3D visual grounding. Conventional methods are difficult to meet the requirements of real-time inference due to the two-stage or point-based architecture. Inspired by the success of multi-level fully sparse convolutional architecture in 3D object detection, we aim to build a new 3D visual grounding framework following this technical route.
However, as in 3D visual grounding task the 3D scene representation should be deeply interacted with text features, sparse convolution-based architecture is inefficient for this interaction due to the large amount of voxel features. 
% To revise, + specifically
To this end, we propose text-guided pruning (TGP) and completion-based addition (CBA) to deeply fuse 3D scene representation and text features in an efficient way by gradual region pruning and target completion.
Specifically, TGP iteratively sparsifies the 3D scene representation and thus efficiently interacts the voxel features with text features by cross-attention. To mitigate the affect of pruning on delicate geometric information, CBA adaptively fixes the over-pruned region by voxel completion with negligible computational overhead.
Compared with previous single-stage methods, our method achieves top inference speed and surpasses previous fastest method by 100\% FPS. Our method also achieves state-of-the-art accuracy even compared with two-stage methods, with $+1.13$ lead of Acc@0.5 on ScanRefer, and $+2.6$ and $+3.2$ leads on NR3D and SR3D respectively. The code is available at \href{https://github.com/GWxuan/TSP3D}{https://github.com/GWxuan/TSP3D}.
\end{abstract}

%% file: sec/1_intro.tex
\section{Introduction}
\label{sec:intro}

Incorporating multi-modal information to guide 3D visual perception is a promising direction. In these years, 3D visual grounding (3DVG), also known as 3D instance referencing, has been paid increasing attention as a fundamental multi-modal 3D perception task. The aim of 3DVG is to locate an object in the scene with a free-form query description. 3DVG is challenging since it requires understanding of both 3D scene and language description. Recently, with the development of 3D scene perception and vision-language models, 3DVG methods have shown remarkable progress~\citep{jain2022bottom,luo20223d}. However, with 3DVG being widely applied in fields like robotics and AR / VR where inference speed is the main bottleneck, how to construct efficient real-time 3DVG model remains a challenging problem. 
% To the best of our knowledge, we are the first to comprehensively study the problem of efficiency for 3DVG.

\begin{figure}
% \vspace{-.3cm}
  \centering
  \includegraphics[width=0.47\textwidth]{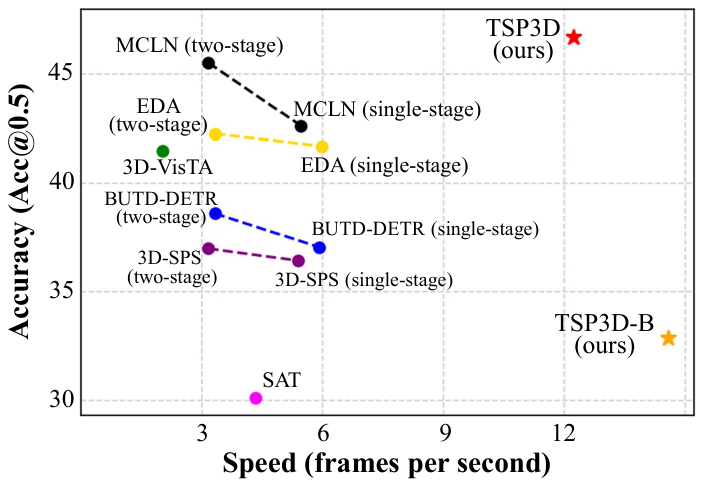}
  \vspace{-.2cm}
  \caption{Comparison of 3DVG methods on ScanRefer dataset~\citep{chen2020scanrefer}. Our TSP3D surpasses existing methods in both accuracy and inference speed, achieving the first efficient 3DVG framework.}
  \label{fig:your_label}
  \vspace{-.3cm}
\end{figure}

Since the output format of 3DVG is similar with 3D object detection, early 3DVG methods~\citep{yuan2021instancerefer,yang2021sat,chen2020scanrefer,
huang2021text} usually adopt a two-stage framework, which first conducts detection to locate all objects in the scene, and then selects the target object by incorporating text information. As there are many similarities between 3D object detection and 3DVG (e.g.\ both of them need to extract the representation of the 3D scene), there will be much redundant feature computation during the independent adoption of the two models. As a result, two-stage methods are usually hard to handle real-time tasks. To solve this problem, single-stage methods~\citep{luo20223d,wu2023eda} are presented, which generates the bounding box of the target directly from point clouds. This integrated design is more compact and efficient. However, current single-stage 3DVG methods mainly build on point-based architecture~\citep{qi2017pointnetpp}, where the feature extraction contains time-consuming operations like furthest point sampling and kNN. They also need to aggressively downsample the point features to reduce computational cost, which might hurt the geometric information of small and thin objects~\citep{xu2023dsp}. Due to these reasons, current single-stage methods are still far from real-time ($<6$ FPS) and their performance is inferior to two-stage methods, as shown in Fig.~\ref{fig:your_label}.

In this paper, we propose a new single-stage framework for 3DVG based on \textbf{t}ext-guided \textbf{s}parse voxel \textbf{p}runing, namely TSP3D. Inspired by state-of-the-art 3D object detection methods~\citep{rukhovich2022fcaf3d,xu2023dsp} which achieves both leading accuracy and speed with multi-level sparse convolutional architecture, we build the first sparse single-stage 3DVG network. However, different from 3D object detection, in 3DVG the 3D scene representation should be deeply interacted with text features. Since the count of voxels is very large in sparse convolution-based architecture, deep multi-modal interaction like cross-attention becomes infeasible due to unaffordable computational cost. To this end, we propose text-guided pruning (TGP), which first utilize text information to jointly sparsify the 3D scene representation and enhance the voxel and text features. To mitigate the affect of pruning on delicate geometric information, we further present completion-based addition (CBA) to adaptively fix the over-pruned region with negligible computational overhead.
Specifically, TGP prunes the voxel features according to the object distribution. It gradually removes background features and features of irrelevant objects, which generates text-aware voxel features around the target object for accurate bounding box prediction. Since pruning may mistakenly remove the representation of target object, CBA utilizes text features to query a small set of voxel features from the complete backbone features, followed by pruned-aware addition to fix the over-pruned region.
We conduct extensive experiments on the popular ScanRefer~\citep{chen2020scanrefer} and ReferIt3D~\citep{achlioptas2020referit3d} datasets.
Compared with previous single-stage methods, TSP3D achieves top inference speed and surpasses previous fastest single-stage method by 100\% FPS. TSP3D also achieves state-of-the-art accuracy even compared with two-stage methods, with $+1.13$ lead of Acc@0.5 on ScanRefer, and $+2.6$ and $+3.2$ leads on NR3D and SR3D respectively.

To summarize, our main contributions are as follows:
\begin{itemize}
	\item[$\bullet$] To the best of our knowledge, this is the first work exploring sparse convolutional architecture for efficient 3DVG.
	\item[$\bullet$] To enable efficient feature extraction, we propose text-guided pruning and completion-based addition to sparsify sparse voxels and adaptively fuse multi-level features.
	\item[$\bullet$] We conduct extensive experiments, and TSP3D outperforms existing methods in both accuracy and speed, demonstrating the superiority of the proposed framework.
\end{itemize}

%% file: sec/2_related.tex
\section{Related Work}
\label{sec:related}

\subsection{3D Visual Grounding}
3D visual grounding aims to locate a target object within a 3D scene based on natural language descriptions~\citep{liu2024surveytextguided3dvisual}. Existing methods are typically categorized into two-stage and single-stage approaches.
Two-stage methods follow a detect-then-match paradigm. In the first stage, they independently extract features from the language query using pre-trained language models~\citep{devlin2018bert,pennington2014glove,chung2014empirical} and predict candidate 3D objects using pre-trained 3D detectors~\citep{qi2019deep,liu2021group} or segmenters~\citep{chen2021hierarchical,jiang2020pointgroup,vu2022softgroup}. In the second stage, they focus on aligning the vision and text features to identify the target object. Techniques for feature fusion include attention mechanisms with Transformers~\citep{he2021transrefer3d,zhao20213dvg}, contrastive learning~\citep{abdelreheem20223dreftransformer}, and graph-based matching~\citep{feng2021free,huang2021text,yuan2021instancerefer}. 
% However, these methods suffer from a detection bottleneck: objects missed in the first stage cannot be recovered later, and they often have high computational overhead.
In contrast, single-stage methods integrate object detection and feature extraction, allowing for direct identification of the target object. Methods in this category include guiding keypoint selection using textual features~\citep{luo20223d}, and measuring similarity between words and objects inspired by 2D image-language pre-trained models like GLIP~\citep{li2022grounded}, as in BUTD-DETR~\citep{jain2022bottom}. And methods like EDA~\citep{wu2023eda} and G$^3$-LQ~\citep{wang2024g} advance single-stage 3D visual grounding by enhancing multimodal feature discriminability through explicit text-decoupling, dense alignment, and semantic-geometric modeling. MCLN~\citep{qian2025multi} uses the 3D referring expression segmentation task to assist 3DVG in improving performance.
% Despite their advantages, existing single-stage methods often have limitations such as reliance on ground-truth annotations for object names, which restricts generalizability, and focusing on sparse alignment of main object words, neglecting dense alignment of all object-related textual components [11].

However, existing two-stage and single-stage methods generally have high computational costs, hindering real-time applications. Our work aims to address these efficiency challenges by proposing an efficient single-stage method with multi-level sparse convolutional architecture.

\subsection{Multi-Level Convolutional Architectures}
Recently, sparse convolutional architecture has achieved great success in the field of 3D object detection. Built on the voxel-based representation~\citep{wang2022cagroup3d,chen2023voxelnext,deng2021voxel} and sparse convolution operation~\citep{choy20194d,graham20183d,xu2023binarizing}, this kind of methods show great efficiency and accuracy when processing scene-level data. GSDN~\citep{gwak2020generative} first adopts multi-level sparse convolution with generative feature upsampling in 3D object detection. FCAF3D~\citep{rukhovich2022fcaf3d} simplifies the multi-level architecture with anchor-free design, achieving leading accuracy and speed. TR3D~\citep{rukhovich2023tr3d} further accelerates FCAF3D by removing unnecessary layers and introducing category-aware proposal assignment method. Moreover, DSPDet3D~\cite{xu2023dsp} introduces the multi-level architecture to 3D small object detection.
% and demonstrates great accuracy and efficiency, even being able to process building-level 3D scenes.

Our proposed method draws inspiration from these approaches, utilizing a sparse multi-level architecture with sparse convolutions and an anchor-free design. This allows for efficient processing of 3D data, enabling real-time performance in 3D visual grounding tasks.

% Anchor-free approaches have demonstrated that voxel-based methods can achieve competitive accuracy without relying on anchors, improving efficiency. FCAF3D [4] is one such method that processes voxelized point clouds through a fully convolutional architecture and employs multi-level feature fusion to enhance detection performance. Similarly, TR3D [25] builds upon FCAF3D and demonstrates that anchor-free, voxel-based methods can achieve competitive results in 3D object detection.

% Our proposed method draws inspiration from these approaches, utilizing a 3DCNN-based multi-level architecture with sparse convolutions and an anchor-free design. This allows for efficient processing of 3D data, enabling real-time performance in 3D visual grounding tasks.

%% file: sec/3_method.tex
\section{Method}
\label{sec:method}

In this section, we describe our TSP3D for efficient single-stage 3DVG. We first analyze existing pipelines to identify current challenges and motivate our approach (Sec. \ref{sec:3.1}). 
% Next, we provide an overview of our framework (Sec. \ref{sec:3.2}). 
We then introduce the text-guided pruning, which leverages text features to guide feature pruning  (Sec. \ref{sec:3.2}). To address the potential risk of pruning key information, we propose the completion-based addition for multi-level feature fusion (Sec. \ref{sec:3.3}). Finally, we detail the training loss (Sec. \ref{sec:3.4}).

\subsection{Architecture Analysis for 3DVG}\label{sec:3.1}

Top-performance 3DVG methods~\citep{wang2024g,wu2023eda,shi2024aware}, are mainly two-stage, which is a serial combination of 3D object detection and 3D object grounding. This separate calls of two approaches result in redundant feature extraction and complex pipeline, thus making the two-stage methods less efficient. 
To demonstrate the efficiency of existing methods, we conduct a comparison of accuracy and speed among several representative methods on ScanRefer~\citep{chen2020scanrefer}, as shown in Fig.~\ref{fig:your_label}. 
% Recognizing that previous studies have not considered computational efficiency, we replicate these methods and report their inference speeds. 
It can be seen that two-stage methods struggle in speed ($<3$ FPS) due to the additional detection stage. Since 3D visual grounding is usually adopted in practical scenarios that require real-time inference under limited resources, such as embodied robots and VR/AR, the low speed of two-stage methods make them less practical.
On the other side, single-stage methods~\citep{luo20223d}, which directly predicts refered bounding box from the observed 3D scene, are more suitable choices due to their streamlined processes. In Fig.~\ref{fig:your_label}, it can be observed that single-stage methods are significantly more efficient than their two-stage counterparts.
% We also list the accuracy-speed tradeoff of single-stage methods in Fig.~\ref{fig:your_label}. It is shown that they are much more efficient than the two-stage counterparts.

However, existing single-stage methods are mainly built on point-based backbone~\citep{qi2017pointnetpp}, where the scene representation is extracted with time-consuming operations like furthest point sampling and set abstraction. They also employ large transformer decoder to fuse text and 3D features for several iterations. Therefore, the inference speed of current single-stage methods is still far from real-time ($<6$ FPS).
The inference speed of specific components in different frameworks is analyzed and discussed in detail in the supplementary material.
Inspired by the success of multi-level sparse convolutional architecture in 3D object detection~\citep{rukhovich2023tr3d}, which achieves both leading accuracy and speed, we propose to build the first multi-level convolutional single-stage 3DVG pipeline.

\begin{figure*}
	\centering
	\includegraphics[width=0.95\linewidth]{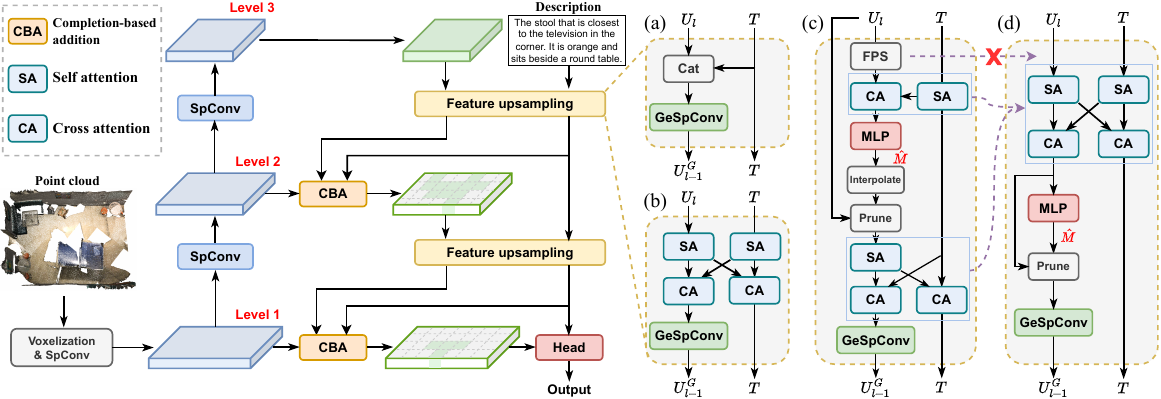}
    \vspace{-.1cm}
    \caption{Illustration of TSP3D. TSP3D bulids on multi-level sparse convolutional architecture. It iteratively upsamples the voxel features with text-guided pruning (TGP), and fuses multi-level features via completion-based addition (CBA). 
    (a) to (d) on the right side illustrate various options for feature upsampling. 
    (a) refers to simple concatenation with text features, which is fast but less accurate. 
    (b) refers to feature interaction through cross-modal attention mechanisms, which is constrained by the large number of voxels. 
    (c) represents our proposed TGP, which first prunes voxel features under textual guidance and thus enables efficient interaction between voxel and text features. 
    (d) shows a simplified version of TGP that removes farthest point sampling and interpolation, combines multi-modal feature interactions into a whole and moves it before pruning.}
	\label{fig:method}
	\vspace{-.2cm}
\end{figure*}

\textbf{TSP3D-B.} Here we propose a baseline framework based on sparse convolution, namely TSP3D-B. Following the simple and effective multi-level architecture of FCAF3D~\citep{rukhovich2022fcaf3d}, TSP3D-B utilizes 3 levels of sparse convolutional blocks for scene representation extraction and bounding box prediction, as shown in Fig.~\ref{fig:method} (a). 
Specifically, the input pointclouds $P\in \mathbb{R}^{N \times 6}$ with 6-dim features (3D position and RGB) are first voxelized and then fed into three sequential MinkResBlocks~\citep{choy20194d}, which generates three levels of voxel features $V_l\ (l=1,2,3)$. With the increase of $l$, the spatial resolution of $V_l$ decreases and the context information increases.
Concurrently, the free-form text with $l$ words is encoded by the pre-trained RoBERTa~\citep{liu2019roberta} and produce the vanilla text tokens \( T  \in \mathbb{R}^{l \times d}\).
With the extracted 3D and text representations, we iteratively upsample $V_3$ and fuse it with $T$ to generate high-resolution and text-aware scene representation:
\begin{equation}\label{eq1}
    U_l=U^G_l+V_l,\ \ \ \ U^G_l={\rm GeSpConv}(U'_{l+1})
\end{equation}
\begin{equation}
    U'_{l+1}={\rm Concat}(U_{l+1}, T) 
\end{equation}
where $U_3=V_3$, ${\rm GeSpConv}$ means generative sparse convolution~\citep{gwak2020generative} with stride 2, which upsamples the voxel features and expands their spatial locations for better bounding box prediction. ${\rm Concat}$ is voxel-wise feature concatenation by duplicating $T$.
The final upsampled feature map $U_1$ is concatenated with $T$ and fed into a convolutional head to predict the objectness scores and regress the 3D bounding box. We select the box with highest objectness score as the grounding result.

As shown in Fig.~\ref{fig:your_label}, TSP3D-B achieves an inference speed of 14.58 FPS, which is significantly faster than previous single-stage methods and demonstrates great potential for real-time 3DVG.

\subsection{Text-guided Pruning}\label{sec:3.2}
Though efficient, TSP3D-B exhibits poor performance due to the inadequate interaction between 3D scene representation and text features. Motivated by previous 3DVG methods~\citep{jain2022bottom}, a simple solution is to replace ${\rm Concat}$ with cross-modal attention to process voxel and text features, as shown in Fig.~\ref{fig:method} (b).
However, different from point-based architectures where the scene representation is usually aggressively downsampled, the number of voxels in multi-level convolutional framework is very large\footnote{Compared to point-based architectures, sparse convolutional framework provides higher resolution and more detailed scene representations, while also offering advantages in inference speed. For detailed statistics, please refer to the supplementary material.}. In practical implementation, we find that the voxels expand almost exponentially with each upsampling layer, leading to a substantial computational burden for the self-attention and cross-attention of scene features. To address this issue, we introduce text-guided pruning (TGP) to construct TSP3D, as illustrated in Fig.~\ref{fig:method} (c). The core idea of TGP is to reduce feature amount by pruning redundant voxels and guide the network to gradually focus on the final target based on textual features.

\textbf{Overall Architecture.} TGP can be regarded as a modified version of cross-modal attention, which reduces the number of voxels before attention operation, thereby lowering computational cost. To minimize the affect of pruning on the final prediction, we propose to prune the scene representation gradually. At higher level where the number of voxels is not too large yet, TGP prunes less voxels. While at lower level where the number of voxels is significantly increased by upsampling operation, TGP prunes the voxel features more aggressively.
The multi-level architecture of TSP3D consists of three levels and includes two feature upsampling operations. Therefore, we correspondingly configure two TGPs with different functions, which are referred as scene-level TGP (level 3 to 2) and target-level TGP (level 2 to 1) respectively. Scene-level TGP aims to distinguish between objects and the background, specifically pruning the voxels on background. Target-level TGP focuses on regions mentioned in the text, intending to preserve the target object and referential objects while removing other regions.

\textbf{Details of TGP.} Since the pruning is relevant to the description, we need to make the voxel features text-aware to predict a proper pruning mask.
To reduce the computational cost, we perform farthest point sampling (FPS) on the voxel features to reduce their size while preserving the basic distribution of the scene. Next, we utilize cross-attention to interact with the text features and employ a simple MLP to predict the probability distribution $\hat{M}$ for retaining each voxel. To prune the features $U_l$, we binarize and interpolate the $\hat{M}$ to obtain the pruned mask. This process can be expressed as:
\begin{equation}\label{eq2} 
    U^P_l = U_l \odot \Theta(\mathcal{I}(\hat{M}, U_l) - \sigma)
\end{equation}
\begin{equation}
    \hat{M} = \text{MLP}(\text{CrossAtt}(\text{FPS}(U_l),\text{SelfAtt}(T)))
\end{equation}
where $U^P_l$ is the pruned features, \(\Theta\) is Heaviside step function, \(\odot\) is matrix dot product, $\sigma$ is the pruning threshold, and \(\mathcal{I}\) represents linear interpolation based on the positions specified by $U_l$. After pruning, the scale of the scene features is significantly reduced, enabling internal feature interactions based on self-attention. Subsequently, we utilize self-attention and cross-attention to perceive the relative relationships among objects within the scene and to fuse multimodal features, resulting in updated features $U'_l$. Finally, through generative sparse convolutions, we obtain $U^G_{l-1}$.

\textbf{Supervision for Pruning.}
The binary supervision mask $M^{sce}$ for scene-level TGP is generated based on the centers of all objects in the scene, and the mask $M^{tar}$ for target-level TGP is based on the target and relevant objects mentioned in the descriptions:
\begin{equation}\label{eq4}
M^{sce} = \bigcup_{i=1}^{N} \mathcal{M}(O_i), \ \
M^{tar} = \mathcal{M}(O^{tar}) \cup \bigcup_{j=1}^{K} \mathcal{M}(O^{rel}_j)
\end{equation}
where $\{O_i|1\leq i\leq N\}$ indicates all objects in the scene. $O^{tar}$ and $O^{rel}$ refer to target and relevant objects respectively.
$\mathcal{M}(O)$ represents the mask generated from the center of object $O$. It generates a \( L \times L \times L \) cube centered at the center of $O$ to construct the supervision mask $M$, where locations inside the cube is set to 1 while others set to 0.

% Our assignment approach does not rely on the dimensions of the bounding box, ensuring a sufficient number of positive proposals for objects of any size.

% replace FPS. 2xinteraction-->1xinteration before pruning.
\textbf{Simplification.} Although the above mentioned method can effectively prune voxel features to reduce the computational cost of cross-modal attention, there are some inefficient operations in the pipeline: (1) FPS is time-consuming, especially for large scenes; (2) there are two times of interactions between voxel features and text features, the first is to guide pruning and the second is to enhance the representation, which is a bit redundant.
We also empirically observe that the number of voxels is not large in level 3. To this end, we propose a simplified version of TGP, as shown in Fig.~\ref{fig:method} (d). We remove the FPS and merge the two multi-modal interactions into one. We also move the merged interaction operation before pruning. In this way, voxel features and text features are first deeply interacted for both feature enhancement and pruning. Because in level 3 the number of voxels is small and in level 2 / 1 the voxels are already pruned, the computational cost of self-attention and cross-attention is always kept at a relatively low level. 

\textbf{Effectiveness of TGP.} After pruning, the voxel count of $U_1$ is reduced to nearly 7\% of its original size without TGP, while the 3DVG performance is significantly boosted.
TGP serves multiple functions, including: (1) facilitating the interaction of multi-modal features through cross-attention, (2) reducing the feature amount (number of voxels) through pruning, and (3) gradually guiding the network to focus on the mentioned target based on text features.

\begin{figure}[t]
  \centering
  \begin{minipage}{0.47\textwidth}
    \centering
    \includegraphics[width=0.88\textwidth]{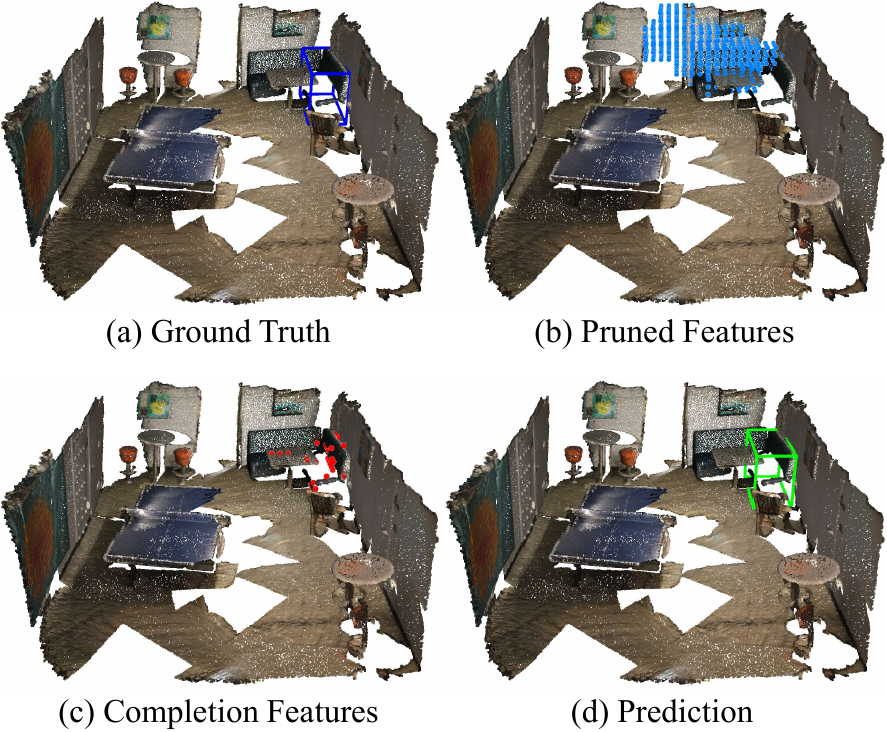}
    % 可以添加对第一张图的简要说明
  \end{minipage}

  \vspace{0.2cm} % 调整上下图之间的间距

  \begin{minipage}{0.47\textwidth}
    \centering
    \includegraphics[width=0.95\textwidth]{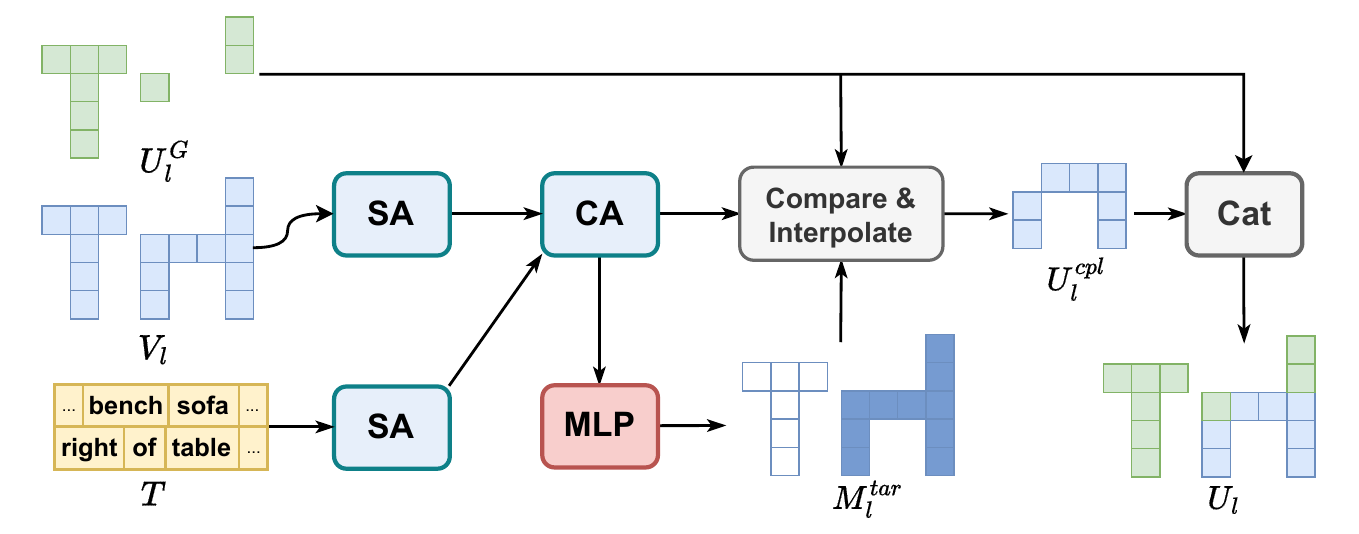}
    % 可以添加对第二张图的简要说明
  \end{minipage}
 \vspace{-0.2cm}
  \caption{Illustration of completion-based addition. The upper figure (b) illustrates an example of over-pruning on the target. The lower figure (c) shows the completed features predicted by CBA.}
  \label{fig:cba}
  \vspace{-0.2cm}
\end{figure}

\subsection{Completion-based Addition}
\label{sec:3.3}
During the pruning process, some targets may be mistakenly removed, especially for small or narrow objects, as shown in Fig.~\ref{fig:cba} (b). Therefore, the addition operation between the upsampled pruned features $U^G_l$ and backbone features $V_l$ described in Equation (\ref{eq1}) play an important role to mitigate the affect of over-pruning.

There are two alternative addition operation: (1) \textbf{Full Addition.} For the intersecting regions of $V_l$ and $U^G_l$, features are directly added. For voxel features outside the intersection of $U^G_l$ and $V_l$ which lack corresponding features in the other map, the missing voxel features are interpolated before addition. Due to pruning process, $U^G_l$ is sparser than $V_l$. In this way, full addition can fix almost all the pruned region. But this operation is computationally heavy and make the scene representation fail to focus on relevant objects, which deviates the core idea of TGP.
(2) \textbf{Pruning-aware Addition.} The addition is constrained to the locations of $U^G_l$. For voxel in $U^G_l$ but not in $V_l$, interpolation from $U^G_l$ is applied to complete the missing locations in $V_l$. It restricts the addition operation to the shape of the pruned features, potentially leading to an over-reliance on the results of the pruning process. If important regions are over-pruned, the network may struggle to detect targets with severely damaged geometric information.

% \begin{figure}{0.5\textwidth}
%  % \vspace{-.8em}
% % \vspace{-.3cm}
%   \centering
%   \begin{minipage}{0.45\textwidth}
%   \centering
%     \includegraphics[width=0.88\textwidth]{fig/method2-1.pdf}
%     % 不添加小 caption
%     % \vspace{.0002em}
%   \end{minipage}
%   % \vspace{5.5em} % 可调节上下间距
%   \begin{minipage}{0.45\textwidth}
%   \centering
%     \includegraphics[width=0.88\textwidth]{fig/method2-2.pdf} % 替换为第二张图的路径
%     % 不添加小 caption
%   \end{minipage}
%   % \vspace{-.5em}
%   \caption{Illustration of completion-based addition. The above (b) illustrate an example of over-pruning on the target. (c) refers to the completed features predicted by CBA. The lower diagram demonstrates how CBA predicts target distribution under textual guidance and adaptively completes the pruned features.}
%   \label{fig:cba}
%   % \vspace{1.5em}
% \end{figure}

Considering the unavoidable risk of pruning the query target, we introduce the completion-based addition (CBA). CBA is designed to address the limitations of full and pruning-aware additions. It offers a more targeted and efficient way to integrating multi-level features, ensuring the preservation of essential details while keeping the additional computational overhead negligible.
% CBA is designed to mitigate the drawback of full and pruning-aware additions by providing a more targeted and efficient way of integrating multi-level features, ensuring that essential details are preserved during the feature fusion process while the additional computational overhead is negligible.

\textbf{Details of CBA.} We first enhance the backbone features $V_l$ with the text features $T$ through cross-attention, obtaining $V_l'$. Then a MLP is adopted to predict the probability distribution of target for region selection:
\begin{equation}
    M_l^{tar} = \Theta(\text{MLP}(V_l') - \tau)
\end{equation}
where \( \Theta \) is the step function, and \( \tau \) is the threshold determining voxel relevance. $M_l^{tar}$ is a binary mask indicating potential regions of the mentioned target.
Then, comparison of $M_l^{tar}$ with $U_l$ identifies missing voxels. The missing mask $M_l^{mis}$ is derived as follows:
\begin{equation}
    M_l^{mis}=M_l^{tar} \land (\neg \ \mathcal{C}(U_l^G,V_l))
\end{equation}
where \(\mathcal{C}(A,B)\) denotes the generation of a binary mask for \(A\) based on the shape of \(B\). Specifically, for positions in \(B\), if there are corresponding voxel features in \(A\), the mask for that position is set to 1. Otherwise it is set to 0.
Missed voxel features in $U_l^G$ that correspond to $M_l^{mis}$ are interpolated from $U_l^G$, filling in gaps identified by the missing mask. The completed feature map $U_l^{cpl}$ is computed by:
\begin{equation}
    U_l^{cpl}=V_l' \odot M_l^{mis} + \mathcal{I}(U_l^G, M_l^{mis})
\end{equation}
where \(\mathcal{I}\) represents linear interpolation on the feature map based on the positions specified in the mask.
Finally, the original upsampled features are combined with the backbone features according to the pruning-aware addition, and merged with the completion features to yield updated $U_l$:
\begin{equation}
    U_l=\text{Concat}(U^G_l \leftarrow V_l, U_l^{cpl})
\end{equation}
where \( \leftarrow \) denotes the pruning-aware addition, and \(\text{Concat}\) means concatenation of voxel features.

% \textbf{Effectiveness of CBA.} CBA is a compromise between adding all potentially over-pruned features and focusing solely on pruned features, ensuring that the final feature map is both comprehensive and sparse, which improves the robustness and efficiency of TSP3D.

\subsection{Train Loss}\label{sec:3.4}

The loss is composed of several components: pruning loss for TGP, completion loss for CBA, and objectness loss as well as bounding box regression loss for the head. Pruning loss, completion loss and objectness loss employ the focal loss to handle class imbalance. 
Supervision for completion and classification losses are the same, which sets voxels near the target object center as positives while leaving others as negatives. 
For bounding box regression, we use the Distance-IoU (DIoU) loss. The total loss function is computetd as the sum of these individual losses:
\[
\mathcal{L}_{\text{total}} = \lambda_1\mathcal{L}_{\text{pruning}} + \lambda_2\mathcal{L}_{\text{com}} + \lambda_3\mathcal{L}_{\text{class}} + \lambda_4\mathcal{L}_{\text{bbox}}
\]
where \(\lambda_1\), \(\lambda_2\), \(\lambda_3\) and \(\lambda_4\) are the weights of different parts.

%% file: sec/4_exper.tex
\section{Experiments}
\label{sec:expre}

\begin{table}[t]
\centering
\caption{Comparison of methods on the ScanRefer dataset evaluated at IoU thresholds of 0.25 and 0.5. TSP3D achieves state-of-the-art accuracy even compared with two-stage methods, with $+1.13$ lead on Acc@0.5. Notably, we are the first to comprehensively evaluate inference speed for 3DVG methods. The inference speeds of other methods are obtained through our reproduction.}
\vspace{-.2cm}
\label{tab:scanrefer}
\footnotesize
\resizebox{0.4777\textwidth}{!}{
\begin{tabular}{@{}cccccc@{}}
\toprule
\multirow{2}{*}{\textbf{Method}} & \multirow{2}{*}{\textbf{Venue}}  & \multirow{2}{*}{\textbf{Input}} & \multicolumn{2}{c}{\textbf{Accuracy}} & \textbf{Inference} \\
% \cmidrule(r){5-6}
~& ~& ~& \textbf{0.25} & \textbf{0.5} & \textbf{Speed (FPS)}\\
\midrule
\multicolumn{6}{l}{\textbf{\textit{Two-Stage Model} }} \\
\midrule
ScanRefer~\citep{chen2020scanrefer} & ECCV'20   & 3D+2D & 41.19 & 27.40 & \textbf{6.72} \\
TGNN~\citep{huang2021text} & AAAI'21   & 3D    & 37.37 & 29.70 & 3.19 \\
InstanceRefer~\citep{yuan2021instancerefer}   & ICCV'21  & 3D    & 40.23 & 30.15 & 2.33 \\
SAT~\citep{yang2021sat} & ICCV'21  & 3D+2D & 44.54 & 30.14 & \underline{4.34} \\
FFL-3DOG~\citep{feng2021free}& ICCV'21  & 3D    & 41.33 & 34.01 & Not released \\
% 3DVG~\citep{zhao20213dvg}  & ICCV'21  & Two-stage & 3D+2D & 47.57 & 34.68 & --- \\
3D-SPS~\citep{luo20223d} & CVPR'22  & 3D+2D & 48.82 & 36.98 & 3.17 \\
BUTD-DETR~\citep{jain2022bottom} & ECCV'22  & 3D & 50.42 & 38.60 & 3.33 \\
% 3D-VLP~\citep{jin2023context} & CVPR'23  & Two-stage & 3D+2D & 51.41 & 39.46 &  \\
EDA~\citep{wu2023eda}  & CVPR'23  & 3D & 54.59 & 42.26 & 3.34 \\
3D-VisTA~\citep{zhu20233d} & ICCV'23  & 3D & 45.90 & 41.50 & 2.03 \\
VPP-Net~\citep{shi2024aware} & CVPR'24  & 3D & 55.65 & 43.29 & Not released \\
\(\text{G}^3\)-LQ~\citep{wang2024g} & CVPR'24  & 3D & \underline{56.90} & \textbf{45.58} & Not released \\
MCLN~\citep{qian2025multi} & ECCV'24  & 3D & \textbf{57.17} & \underline{45.53} & 3.17 \\
% TSP3D(Ours)    & -----  & Two-stage & 3D &   &   &  --- \\
\midrule
\multicolumn{6}{l}{\textbf{\textit{Single-stage Model} }} \\
\midrule
3D-SPS~\citep{luo20223d} & CVPR'22  & 3D & 47.65 & 36.43 & 5.38 \\
BUTD-DETR~\citep{jain2022bottom}  & ECCV'22  & 3D & 49.76 & 37.05 & 5.91 \\
EDA~\citep{wu2023eda}  & CVPR'23  & 3D & 53.83 & 41.70 & \underline{5.98} \\
\(\text{G}^3\)-LQ~\citep{wang2024g} & CVPR'24  & 3D & \underline{55.95} & \underline{44.72} & Not released \\
MCLN~\citep{qian2025multi} & ECCV'24  & 3D & 54.30 & 42.64 & 5.45 \\
TSP3D (Ours)    & -----  & 3D & \textbf{56.45} & \textbf{46.71} &  \textbf{12.43} \\
\bottomrule
\end{tabular}
}
\vspace{-.2cm}
\end{table}

\subsection{Datasets}
We maintain the same experimental settings with previous works, employing ScanRefer~\citep{chen2020scanrefer} and SR3D/NR3D~\citep{achlioptas2020referit3d} as datasets.
\textbf{ScanRefer}: Built on the ScanNet framework, ScanRefer includes 51,583 descriptions across scenes. Evaluation metrics focus on Acc@\textit{m}IoU.
\textbf{ReferIt3D}: ReferIt3D splits into Nr3D, with 41,503 human-generated descriptions, and Sr3D, containing 83,572 synthetic expressions. ReferIt3D simplifies the task by providing segmented point clouds for each object. The primary evaluation metric is accuracy in target object selection.

\subsection{Implementation Details}
TSP3D is implemented based on PyTorch~\citep{paszke2019pytorch}. 
The pruning thresholds are set at \(\sigma_\text{sce} = 0.7\) and \(\sigma_\text{tar} = 0.3\), and the completion threshold in CBA is \(\tau = 0.15 \). The initial voxelization of the point cloud has a voxel size of 1cm, while the voxel size for level \(i\) features scales to \(2^{i+2}\) cm. The supervision for pruning uses \(L = 7\). The weights for all components of the loss function, \(\lambda_1, \lambda_2, \lambda_3, \lambda_4\), are equal to 1. Training is conducted using four GPUs, while inference speeds are evaluated using a single consumer-grade GPU, RTX 3090, with a batch size of 1.

\begin{table}[t]
\centering
\caption{Quantitative comparisons on Nr3D and Sr3D datasets. We evaluate under three pipelines, noting that the Two-stage using Ground-Truth Boxes is impractical for real-world applications. TSP3D exhibits significant superiority, with leads of $+2.6\%$ and $+3.2\%$ on NR3D and SR3D respectively.}
 \vspace{-0.2cm}
\label{tab:comparison_nr3d_sr3d}
\footnotesize
\resizebox{0.47\textwidth}{!}{
\begin{tabular}{ccccccc}
\toprule
\multirow{2}{*}{\textbf{Method}} & \multirow{2}{*}{\textbf{Venue}} & \multirow{2}{*}{\textbf{Pipeline}} & \multicolumn{2}{c}{\textbf{Accuracy}}\\ %\cmidrule(lr){4-5} \cmidrule(lr){6-7}
 ~   &    ~   &     ~     & \textbf{Nr3D} & \textbf{Sr3D} \\ 
\midrule
% TGNN [16]           & ECCV'20 & Two-stage (gt) & 37.3 & 45.0 & --- & --- \\
\rowcolor{gray!20} InstanceRefer~\citep{yuan2021instancerefer}   & ICCV'21 & Two-stage (gt) & 38.8 & 48.0  \\
% 3DVG [48]           & ICCV'21 & Two-stage (gt) & 38.9 & 51.4 & --- & --- \\
\rowcolor{gray!20} LanguageRefer~\citep{roh2022languagerefer} & CoRL'22 & Two-stage (gt) & 43.9 & 56.0  \\
% \rowcolor{gray!20} TransRefer3D [15]   & ECCV'20 & Two-stage (gt) & 48.0 & 57.4 & --- & --- \\
% SAT [44]            & ICCV'21 & Two-stage (gt) & 49.2 & 57.9 & --- & --- \\
% LAR [4]             & CVPR'21 & Two-stage (gt) & 48.9 & 59.4 & --- & --- \\
% 3DRef [2]           & NeurIPS'22 & Two-stage (gt) & 47.0 & 39.0 & --- & --- \\
\rowcolor{gray!20} 3D-SPS~\citep{luo20223d} & CVPR'22 & Two-stage (gt) & 51.5 & 62.6 \\
\rowcolor{gray!20} MVT~\citep{huang2022multi}  & CVPR'22 & Two-stage (gt) & 55.1 & 64.5 \\
\rowcolor{gray!20} BUTD-DETR~\citep{jain2022bottom}  & ECCV'22 & Two-stage (gt) & 54.6 & 67.0 \\
\rowcolor{gray!20} EDA~\citep{wu2023eda}   & CVPR'23 & Two-stage (gt) & 52.1 & 68.1 \\
% \rowcolor{gray!20} 3D-VisTA~\citep{zhu20233d}   & ICCV'23 & Two-stage (gt) & 57.5 & 69.6 \\
\rowcolor{gray!20} VPP-Net~\citep{shi2024aware}  & CVPR'24 & Two-stage (gt) & 56.9 & 68.7 \\
\rowcolor{gray!20} \(\text{G}^3\)-LQ~\citep{wang2024g}  & CVPR'24 & Two-stage (gt) & 58.4 & 73.1 \\
\rowcolor{gray!20} MCLN~\citep{qian2025multi} & ECCV'24 & Two-stage (gt) & 59.8 & 68.4 \\
\midrule
InstanceRefer~\citep{yuan2021instancerefer}  & ICCV'21 & Two-stage (det) & 29.9 & 31.5 \\
LanguageRefer~\citep{roh2022languagerefer}  & CoRL'22 & Two-stage (det) &  28.6 & 39.5 \\
BUTD-DETR~\citep{jain2022bottom} & ECCV'22 & Two-stage (det) & \underline{43.3} & \underline{52.1} \\
EDA~\citep{wu2023eda}   & CVPR'23 & Two-stage (det) & 40.7 & 49.9  \\
MCLN~\citep{qian2025multi} & ECCV'24 & Two-stage (det) & \textbf{46.1} & \textbf{53.9} \\
\midrule
3D-SPS~\citep{luo20223d}  & CVPR'22 & Single-stage & 39.2 & 47.1 \\
BUTD-DETR~\citep{jain2022bottom} & ECCV'22 & Single-stage & 38.7 & 50.1 \\
EDA~\citep{wu2023eda}   & CVPR'23 & Single-stage & 40.0 & 49.7  \\
MCLN~\citep{qian2025multi} & ECCV'24 & Single-stage & \underline{45.7} & \underline{53.4} \\
TSP3D (Ours)    & ----- & Single-stage & \textbf{48.7} & \textbf{57.1} \\
\bottomrule
\end{tabular}
}
\vspace{-.3cm}
\end{table}

\begin{table*}[t]
\centering
\begin{minipage}{0.3\textwidth}
  \caption{Impact of the proposed TGP and CBA. Evaluated on ScanRefer.}
  \vspace{-.2cm}
  \label{tab:ablation1}
  % \footnotesize
  \resizebox{\textwidth}{!}{
  \begin{tabular}{@{}ccc|cccc@{}}
    \toprule
    \multirow{2}{*}{\textbf{ID}} & \multirow{2}{*}{\textbf{TGP}} & \multirow{2}{*}{\textbf{CBA}} & \multicolumn{2}{c}{\textbf{Accuracy}} & \multirow{2}{*}{\textbf{Speed (FPS)}} \\
    ~ & ~ & ~ & \textbf{0.25} & \textbf{0.5} & ~ \\
    \midrule
    (a) &  &            & 40.13 & 32.87 & \textbf{14.58} \\
    (b) & \checkmark &  & 55.20 & 46.15 & 13.22 \\
    (c) &  & \checkmark & 41.34 & 33.09 & 13.51 \\
    (d) & \checkmark & \checkmark & \textbf{56.45} & \textbf{46.71} & 12.43 \\
    \bottomrule
  \end{tabular}
  }
\end{minipage}%
\hfill
\begin{minipage}{0.332\textwidth}
  \caption{Influence of the two CBAs at different levels. Evaluated on ScanRefer.}
  \vspace{-.2cm}
  \label{tab:ablation2}
  % \footnotesize
  \resizebox{\textwidth}{!}{
  \begin{tabular}{@{}ccc|cccc@{}}
    \toprule
    \multirow{2}{*}{\textbf{ID}} & \textbf{CBA} & \textbf{CBA} & \multicolumn{2}{c}{\textbf{Accuracy}} & \multirow{2}{*}{\textbf{Speed (FPS)}} \\
    ~ & \textbf{(level 2)} & \textbf{(level 1)} & \textbf{0.25} & \textbf{0.5} & ~ \\
    \midrule
    (a) &  &            & 55.20 & 46.15 & \textbf{13.22} \\
    (b) & \checkmark &  & 55.17 & 46.06 & 12.79 \\
    (c) &  & \checkmark & \textbf{56.45} & \textbf{46.71} & 12.43 \\
    (d) & \checkmark & \checkmark & 56.22 & 46.68 & 12.19 \\
    \bottomrule
  \end{tabular}
  }
\end{minipage}%
\hfill
\begin{minipage}{0.334\textwidth}
  \caption{Influence of different feature upsampling methods. Evaluated on ScanRefer.}
  \vspace{-.2cm}
  \label{tab:ablation3}
  % \footnotesize
  \resizebox{\textwidth}{!}{
  \begin{tabular}{@{}cc|cccc@{}}
    \toprule
    \multirow{2}{*}{\textbf{ID}} & \multirow{2}{*}{\textbf{Method}} & \multicolumn{2}{c}{\textbf{Accuracy}} & \multirow{2}{*}{\textbf{Speed (FPS)}} \\
    ~ & ~ & \textbf{0.25} & \textbf{0.5} & ~ \\
    \midrule
    (a) & Simple concatenation  & 40.13 & 32.87 & \textbf{14.58} \\
    (b) & Attention mechanism  & --- & --- & --- \\
    (c) & Text-guided pruning  & 56.27 & 46.58 & 10.11 \\
    (d) & Simplified TGP & \textbf{56.45} & \textbf{46.71} & 12.43 \\
    \bottomrule
  \end{tabular}
  }
\end{minipage}
\vspace{-.3cm}
\end{table*}

\subsection{Quantitative Comparisons}
\textbf{Performance on ScanRefer.}
% Our proposed method, \textbf{TSP3D}, has demonstrated state-of-the-art performance on the ScanRefer dataset, as detailed in Tab.~\ref{tab:scanrefer}. In both single and two-stage settings, \textbf{TSP3D} outperforms all competing methods by achieving the highest scores in terms of Acc@0.25 and ACC@0.5.
We carry out comparisons with existing methods on ScanRefer, as detailed in Tab.~\ref{tab:scanrefer}. The inference speeds of other methods are obtained through our reproduction with a single RTX 3090 and a batch size of 1. For two-stage methods, the inference speed includes the time taken for object detection in the first stage. For methods using 2D image features and 3D point clouds as inputs, we do not account for the time spent extracting 2D features, assuming they can be obtained in advance. However, in practical applications, the acquisition of 2D features also impacts overall efficiency.
TSP3D achieves state-of-the-art accuracy even compared with two-stage methods, with $+1.13$ lead on Acc@0.5. Notably, in the single-stage setting, TSP3D achieves fast inference speed, which is unprecedented among the existing methods.
This significant improvement is attributed to our method's efficient use of a multi-level architecture based on 3D sparse convolutions, coupled with the text-guided pruning. By focusing computation only on salient regions of the point clouds, determined by textual cues, our model effectively reduces computational overhead while maintaining high accuracy. 
% This enables our system to provide a viable solution for real-time efficient 3D visual grounding.
TSP3D also sets a benchmark for inference speed comparisons for future methodologies.

\textbf{Performance on Nr3D/Sr3D.}
We evaluate our method on the SR3D and NR3D datasets, following the evaluation protocols of prior works like EDA~\citep{wu2023eda} and BUTD-DETR~\citep{jain2022bottom} by using Acc@0.25 as the accuracy metric. The results are shown in Tab.~\ref{tab:comparison_nr3d_sr3d}.
Given that SR3D and NR3D provide ground-truth boxes and categories for all objects in the scene, we consider three pipelines: (1) Two-stage using Ground-Truth Boxes, (2) Two-stage using Detected Boxes, and (3) Single-stage.
In practical applications, the Two-stage using Ground-Truth Boxes pipeline is unrealistic because obtaining all ground-truth boxes in a scene is infeasible. This approach can also oversimplify certain evaluation scenarios. For example, if there are no other objects of the same category as the target in the scene, the task reduces to relying on the provided ground-truth category.
% Conversely, the Two-stage using Detected Boxes and Single-stage pipelines are more practical and reflective of real-world conditions. 
Under the Single-stage setting, TSP3D exhibits significant superiority with peak performance of $48.7\%$ and $57.1\%$ on Nr3D and Sr3D. 
TSP3D even outperforms previous works under the pipeline of Two-stage using Detected Boxes, with leads of $+2.6\%$ and $+3.2\%$ on NR3D and SR3D.

\subsection{Ablation Study}

\textbf{Effectiveness of Proposed Components.} 
To investigate the effects of our proposed TGP and CBA, we conduct ablation experiments with module removal as shown in Tab.~\ref{tab:ablation1}. When TGP is not used, multi-modal feature concatenation is employed as a replacement, as shown in Fig.~\ref{fig:method} (a). When CBA is not used, it is substituted with a pruning-based addition. The results demonstrate that TGP significantly enhances performance without notably impacting inference time. This is because TGP, while utilizing a more complex multi-modal attention mechanism for stronger feature fusion, significantly reduces feature scale through text-guided pruning. Additionally, the performance improvement is also due to the gradual guidance towards the target object by both scene-level and target-level TGP. Using CBA alone has a limited effect, as no voxels are pruned. Implementing CBA on top of TGP further enhances performance, as CBA dynamically compensates for some of the excessive pruning by TGP, thus increasing the network's robustness.

\textbf{Influence of the Two CBAs.}
To explore the impact of CBAs at two different levels, we conduct ablation experiments as depicted in Tab.~\ref{tab:ablation2}. In the absence of CBA, we use pruning-based addition as a substitute. The results indicate that the CBA at level 2 has negligible effects on the 3DVG task. This is primarily because the CBA at level 2 serves to supplement the scene-level TGP, which is expected to prune the background (a relatively simple task). Moreover, although some target features are pruned, they are compensated by two subsequent generative sparse convolutions. However, the CBA at level 1 enhances performance by adapt completion for the target-level TGP. It is challenging to fully preserve target objects from deep upsampling features, especially for smaller or narrower targets. The CBA at level 1, based on high-resolution backbone features, effectively complements the TGP.

\begin{figure*}[t]
	\centering
	\includegraphics[width=0.85\linewidth]{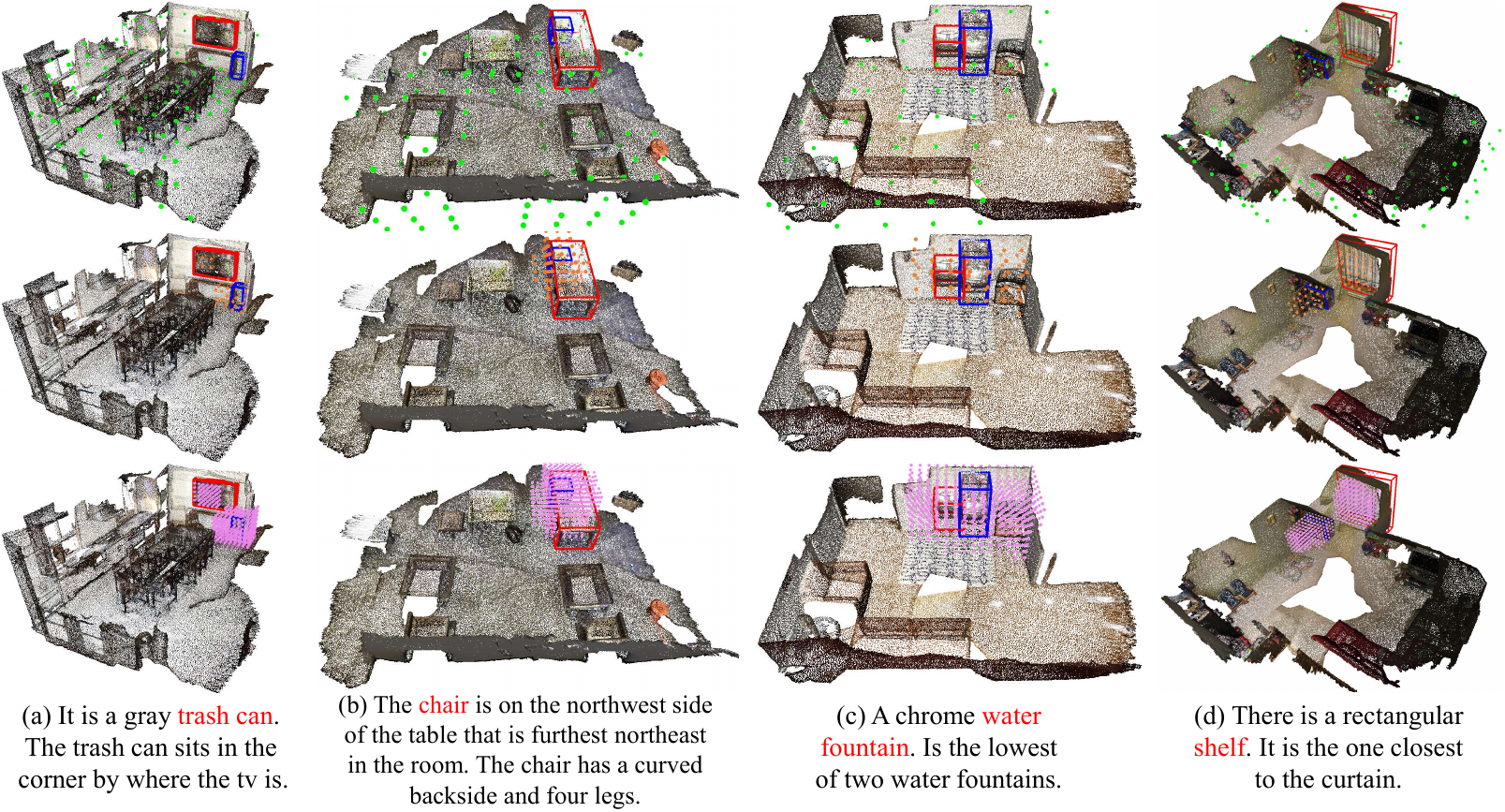}
    \vspace{-.2cm}
    \caption{Visualization of the text-guided pruning process. In each example, the voxel features after scene-level TGP, target-level TGP and the last upsampling layer are presented from top to bottom. The blue boxes represent the ground truth of the target, and the red boxes denote the bounding boxes of relevant objects. TSP3D reduces the amount of voxel features through two stages of pruning and progressively guides the network focusing towards the target.}
	\label{fig:prune}
	\vspace{-.1cm}
\end{figure*}

\begin{figure*}[t]
	\centering
	\includegraphics[width=0.85\linewidth]{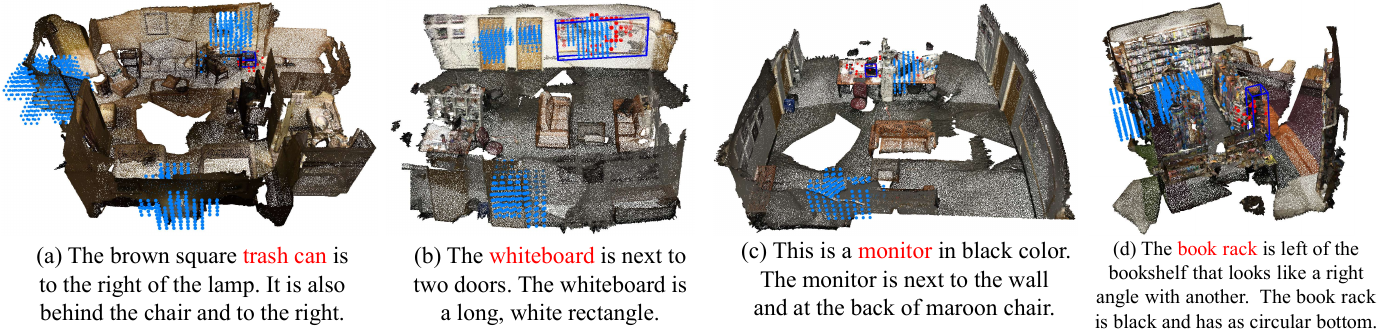}
    \vspace{-.2cm}
    \caption{Visualization of the completion-based addition process. The blue points represent the voxel features output by the target-level TGP, while the red points are the completion features predicted by the CBA. The blue boxes indicate the ground truth boxes. CBA adaptively supplements situations where excessive pruning has occurred.}
	\label{fig:com}
    \vspace{-.1cm}
\end{figure*}

\textbf{Feature Upsampling Techniques.}
We conduct experiments to assess the effects of different feature upsampling techniques, as detailed in Tab.~\ref{tab:ablation3}. Using simple feature concatenation (Fig.~\ref{fig:method} (a)), while fast in inference speed, results in poor performance. When we utilize an attention mechanism with stronger feature interaction, as shown in Fig.~\ref{fig:method} (b), the computation exceeds the limits of GPU due to the large number of voxels, making it impractical for real-world applications. Consequently, we employ TGP to reduce the feature amount, as illustrated in Fig.~\ref{fig:method} (c), which significantly improves performance and enables practical deployment. Building on TGP, we propose simplified TGP, as shown in Fig.~\ref{fig:method} (d), that merges feature interactions before and after pruning, achieving performance consistent with the original TGP while enhancing inference speed.

% \begin{table}[ht]
% \centering
% \caption{Ablation study on the influence of the two CBAs in different levels. Evaluated on the ScanRefer dataset.}
% \vspace{.2cm}
% \label{tab:ablation_study}
% \begin{tabular}{@{}ccc|cccc@{}}
% \toprule
% \multirow{2}{*}{\textbf{ID}} & \textbf{CBA} & \textbf{CBA} & \multicolumn{2}{c}{\textbf{Accuracy}} & \multirow{2}{*}{\textbf{Speed (FPS)}} \\
% % \cmidrule(lr){4-5}
% ~ & \textbf{(level 1)} & \textbf{(level 2)} & \textbf{0.25} & \textbf{0.5} & ~ \\
% \midrule
% (a) &  &            & 55.20 & 46.15 & \textbf{13.22} \\
% (b) & \checkmark &  & 55.17 & 46.06 & 12.79 \\
% (c) &  & \checkmark & \textbf{56.45} & \textbf{46.71} & 12.43 \\
% (d) & \checkmark & \checkmark & 56.22 & 46.68 & 12.19 \\
% \bottomrule
% \end{tabular}
% \end{table}

% \begin{table}[t]
% \centering
% \caption{Influence of different feature upsampling methods. Evaluated on the ScanRefer dataset.}
% \vspace{-.1cm}
% \label{tab:ablation3}
% \footnotesize
% \begin{tabular}{@{}cc|cccc@{}}
% \toprule
% \multirow{2}{*}{\textbf{ID}} & \multirow{2}{*}{\textbf{Method}} & \multicolumn{2}{c}{\textbf{Accuracy}} & \multirow{2}{*}{\textbf{Speed (FPS)}} \\
% % \cmidrule(lr){4-5}
% ~ & ~ & \textbf{0.25} & \textbf{0.5} & ~ \\
% \midrule
% (a) & Simple concatenation  & 40.13 & 32.87 & \textbf{14.58} \\
% (b) & Attention mechanism  & --- & --- & --- \\
% (c) & Text-guided Pruning  & 56.27 & 46.58 & 10.11 \\
% (d) & Simplified TGP & \textbf{56.45} & \textbf{46.71} & 12.43 \\
% \bottomrule
% \end{tabular}
% \end{table}

\subsection{Qualitative Results}
\textbf{Text-guided Pruning.}
To visually demonstrate the process of TGP, we visualize the results of two pruning phases, as shown in Fig.~\ref{fig:prune}. In each example, the voxel features after scene-level pruning, the features after target-level pruning, and the features after target-level generative sparse convolution are displayed from top to bottom. It is evident that both pruning stages effectively achieve our intended effect: the scene-level pruning filters out the background and retained object voxels, and the target-level pruning preserves relevant and target objects. Moreover, during the feature upsampling process, the feature amount nearly exponentially increases due to generative upsampling. Without TGP, the voxel coverage would far exceed the range of the scene point cloud, which is inefficient for inference. This also intuitively explains the significant impact of our TGP on both performance and inference speed.

\textbf{Completion-based Addition.}
To clearly illustrate the function of CBA, we visualize the adaptive completion process in Fig.~\ref{fig:com}. The images below showcase several instances of excessive pruning. TGP performs pruning based on deep and low-resolution features, which can lead to excessive pruning, potentially removing entire or partial targets. This over-pruning is more likely to occur with small, as shown in Fig.~\ref{fig:com} (a) and (c), narrow, as in Fig.~\ref{fig:com} (b), or elongated targets, as in Fig.~\ref{fig:com} (d). Our CBA effectively supplements the process using higher-resolution backbone features, thus dynamically integrating multi-level features.

%% file: sec/5_conclu.tex
\section{Conclusion}
\label{sec:conclu}

In this paper, we present TSP3D, an efficient sparse single-stage method for real-time 3D visual grounding. Different from previous 3D visual grounding frameworks, TSP3D builds on multi-level sparse convolutional architecture for efficient and fine-grained scene representation extraction. To enable the interaction between voxel features and textual features, we propose text-guided pruning (TGP), which reduces the amount of voxel features and guides the network to progressively focus on the target object. Additionally, we introduce completion-based addition (CBA) for adaptive multi-level feature fusion, effectively compensating for instances of over-pruning. Extensive experiments demonstrate the effectiveness of our proposed modules, resulting in an efficient 3DVG method that achieves state-of-the-art accuracy and fast inference speed.

% \textbf{Potential Limitations.} Despite of the leading accuracy and inference speed, there are still some limitations of TSP3D. First, the speed of TSP3D is bit slower than TSP3D-B. Although TSP3D utilizes TGP to enable deep interaction between voxel and text features in an efficient way, it unavoidably introduces additional computational overhead compared with naive concatenation. In the future work, we aim to work on designing new operations for multi-modal feature interaction to replace the heavy cross-attention mechanism. Second, currently the input of 3DVG methods is a reconstructed point clouds. We will work on extending it to online setting with streaming RGB-D videos as input, which can support a wider range of practical application.

%% file: sec/X_suppl.tex
\clearpage
\setcounter{page}{1}
\appendix
\maketitlesupplementary

We provide statistics and analysis for visual feature resolution (Sec.~\ref{sec:supp1}), detailed comparisons of computational cost (Sec.~\ref{sec:supp2}), detailed results on the ScanRefer dataset~\citep{chen2020scanrefer} (Sec.~\ref{sec:supp3}), qualitative comparisons (Sec.~\ref{sec:supp4}) and potential limitations (Sec.~\ref{sec:supp5}) in the supplementary material.

\section{Visual Feature Resolution of Different Architectures}\label{sec:supp1}
\begin{figure}[h]
	\centering
	\includegraphics[width=0.97\linewidth]{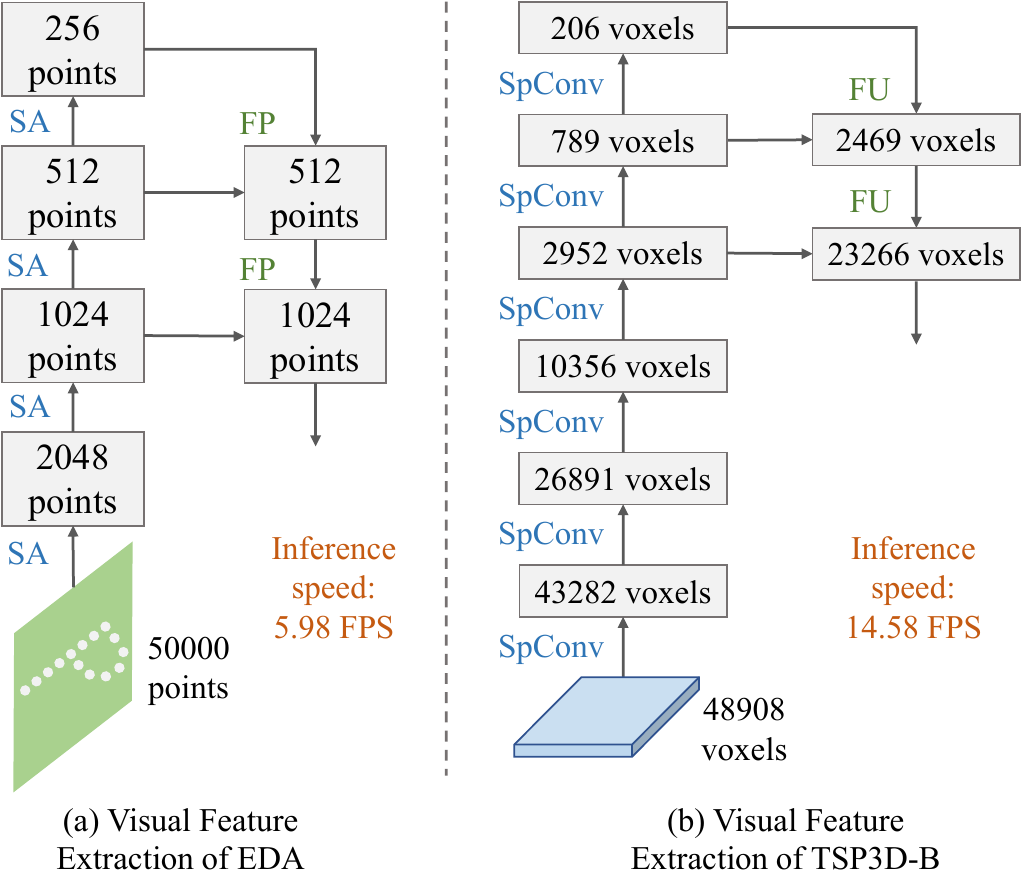}
    \caption{Feature resolution progression of point-based EDA and sparse convolutional TSP3D-B. \textcolor[rgb]{0.2, 0.457, 0.711}{SA}, \textcolor[rgb]{0.328, 0.508, 0.207}{FP}, \textcolor[rgb]{0.2, 0.457, 0.711}{SpConv}, and \textcolor[rgb]{0.328, 0.508, 0.207}{FU} represent set abstraction, feature propagation, sparse convolution, and feature upsampling, respectively. For the point-based architecture, the downsampling process is aggressive, with the first downsampling reducing 50,000 points directly to 2,048 points. Furthermore, the final scene representation consists of only 1,024 points. In contrast, the sparse convolutional architecture performs progressive downsampling and refines the scene representation through a multi-level structure. This approach not only provides a high-resolution scene representation but also achieves faster inference speed compared to the point-based architecture.}
	\label{fig:supp2}
	%\vspace{-.2cm}
\end{figure}

To analyze the scene representation resolution of point-based and sparse convolutional architectures, we compare the resolution changes during the visual feature extraction process for EDA~\citep{wu2023eda} and TSP3D-B, as illustrated in Fig.~\ref{fig:supp2}. For a thorough examination of the feature resolution of the sparse convolution architecture, we consider TSP3D-B without incorporating TGP and CBA.
The voxel numbers for TSP3D-B are based on the average statistics from the ScanRefer validation set. In point-based architectures, the number of point features is fixed and does not vary with the scene size. In contrast, the number of voxel features in sparse convolutional architectures tends to increase as the scene size grows. This adaptive adjustment ensures that features do not become excessively sparse when processing larger scenes. 
As shown in Fig.~\ref{fig:supp2}, point-based architectures perform aggressive downsampling, with the first downsampling step reducing 50,000 points to just 2,048 points. Moreover, the final scene representation consists of only 1,024 points, leading to a relatively coarse representation. By contrast, convolution-based architectures progressively downsample and refine the scene representation through a multi-level structure. Overall, the sparse convolution architecture not only provides high-resolution scene representation but also achieves faster inference speed compared to point-based architectures.

\begin{table*}[h]
\centering
\caption{Detailed comparison of computational cost for different single-stage architectures on the ScanRefer dataset~\citep{chen2020scanrefer}. The numbers in the table represent frames per second (FPS). TSP3D demonstrates superior processing speed across all components compared to other methods, with the inference speed of the sparse convolution backbone being three times faster than that of the point-based backbone.}
 % \vspace{-0.2cm}
\label{tab:detal_speed}
% \setlength{\tabcolsep}{3.75pt}
% \footnotesize
\resizebox{0.8\textwidth}{!}{
\begin{tabular}{ccccccc}
\toprule
\multirow{2}{*}{\textbf{Method}} & \textbf{Text} & \textbf{Visual} & \textbf{Text} &  \textbf{Multi-modal} & \multirow{2}{*}{\textbf{Head}} & \multirow{2}{*}{\textbf{Overall}}\\ 
~ & \textbf{Decouple} & \textbf{Backbone} & \textbf{Backbone} &  \textbf{Fusion} & ~ & ~\\ 
\midrule
3D-SPS~\citep{luo20223d}  & --- & 10.88 & 80.39 & 13.25 & \underline{166.67} & 5.38\\
BUTD-DETR~\citep{jain2022bottom} & 126.58 & 10.60 & 78.55 & 28.49 & 52.63 & 5.91\\
EDA~\citep{wu2023eda}   & 126.58 & \underline{10.89} & \underline{81.10} & \underline{28.57} & 49.75 & \underline{5.98}\\
MCLN~\citep{qian2025multi} & 126.58 & 10.52 & 76.92 & 23.26 & 41.32 & 5.45 \\
% \rowcolor{orange!80} 
TSP3D (Ours)    & --- & \textbf{31.88} & \textbf{81.21} & \textbf{28.67} & \textbf{547.32} & \textbf{12.43} \\
\bottomrule
\end{tabular}
}
% \vspace{-.3cm}
\end{table*}

\section{Detailed Computational Cost of Different Architectures}\label{sec:supp2}
We provide a detailed comparison of the inference speed of specific components across different architectures, as shown in Tab.~\ref{tab:detal_speed}. Two-stage methods tend to have slower inference speed and are significantly impacted by the efficiency of the detection stage, which is not the primary focus of the 3DVG task. Therefore, we focus our analysis solely on the computational cost of single-stage methods.
We divide the networks of existing methods and TSP3D into several components: text decoupling, visual backbone, text backbone, multi-modal fusion, and the head. The inference speed of each of these components is measured separately.

\textbf{Backbone.} Except for TSP3D, the visual backbone in other methods is PointNet++~\citep{qi2017pointnetpp}, which has a high computational cost. This is precisely why we introduce a sparse convolution backbone, which achieves approximately three times the inference speed of PointNet++. As for the text backbone, both TSP3D and other methods use the pre-trained RoBERTa~\citep{liu2019roberta}, so the inference speed for this component is largely consistent across the methods.

\textbf{Multi-modal Fusion.} The multi-modal feature fusion primarily involves the interaction between textual and visual features, with different methods employing different modules. For instance, the multi-modal fusion in SDSPS mainly includes the description-aware keypoint sampling (DKS) and target-oriented progressive mining (TPM) modules. And methods like BUTD-DETR, EDA, and MCLN rely on cross-modal encoders and decoders for their fusion process.
In our TSP3D, the multi-modal fusion involves feature upsampling, text-guided pruning (TGP), and completion-based addition (CBA). Notably, even though TSP3D progressively increases the resolution of scene features and integrates them with fine-grained backbone features, it still achieves superior inference speed. This is primarily due to the text-guided pruning, which significantly reduces the number of voxels and computational cost.

\textbf{Head and Text Decouple.} In the designs of methods such as BUTD-DETR, EDA, and MCLN, the input text needs to be decoupled into several semantic components. 
Additionally, their heads do not output prediction scores directly. Instead, they output embeddings for each candidate object, which must be compared with the embeddings of each word in the text to compute similarities and determine the final output.
This can be considered additional pre-processing and post-processing steps, with the latter significantly impacting computational efficiency.
In contrast, our TSP3D directly predicts the matching scores between the objects and the input text, making the head inference speed over ten times faster than these methods.

% \begin{table}[]
% \centering
% \caption{Detailed comparison of computational cost for different single-stage architectures on the ScanRefer dataset.}
%  % \vspace{-0.2cm}
% \label{tab:comparison_nr3d_sr3d}
% \footnotesize
% \resizebox{0.477\textwidth}{!}{
% \begin{tabular}{ccccccc}
% \toprule
% \textbf{Method} & \textbf{Text Decouple} & \textbf{Visual Backbone} & \textbf{Text Backbone} &  \textbf{Multi-modal Fusion} & \textbf{Head} & \textbf{Overall}\\ 
% \midrule
% 3D-SPS~\citep{luo20223d}  & 22.0 & 22.0 & 22.0 & 22.0 \\
% BUTD-DETR~\citep{jain2022bottom} & 22.0 & 22.0 & 22.0 & 22.0 \\
% EDA~\citep{wu2023eda}   & 22.0 & 22.0 & 22.0 & 22.0  \\
% MCLN~\citep{qian2025multi} & 22.0 & 22.0 & 22.0 & 22.0 \\
% TSP3D (Ours)    & --- & 22.0 & 22.0 & 22.0 \\
% \bottomrule
% \end{tabular}
% }
% % \vspace{-.3cm}
% \end{table}

\section{Detailed Results on ScanRefer}\label{sec:supp3}
Due to page limitations, we report only the overall performances and inference speeds in the main text. To provide detailed results and analysis, we include the accuracies of TSP3D and other methods across various subsets on the ScanRefer dataset~\citep{chen2020scanrefer}, as shown in Tab.~\ref{tab:supp}. TSP3D achieves state-of-the-art accuracy, even when compared with two-stage methods, leading by $+1.13$ in Acc@0.5. TSP3D also demonstrates a level of efficiency that previous methods lack. In various subsets, TSP3D maintains comparable accuracy to both single-stage and two-stage state-of-the-art methods.
Notably, the ``multi-object" subset involves distinguishing the target object among numerous distractors of the same category within a more complex 3D scene. In this setting, TSP3D achieves a commendable performance of \(42.37\) in Acc@0.5, further demonstrating that TSP3D enhances attention to the target object in complex environments through text-guided pruning and completion-based addition, enabling accurate predictions of both the location and the shape of the target.

\begin{table*}[h]
\centering
\caption{Detailed comparison of methods on the ScanRefer dataset~\citep{chen2020scanrefer} evaluated at IoU thresholds of 0.25 and 0.5. TSP3D achieves state-of-the-art accuracy even compared with two-stage methods, with $+1.13$ lead on Acc@0.5. In various subsets, TSP3D achieves comparable accuracy to both single-stage and two-stage state-of-the-art methods. Additionally, TSP3D demonstrates a level of efficiency that previous methods lack.}
% \vspace{.2cm}
\label{tab:supp}
% \footnotesize
\resizebox{0.95\textwidth}{!}{
\begin{tabular}{@{}ccccccccc@{}}
\toprule
\multirow{2}{*}{\textbf{Method}} & \multirow{2}{*}{\textbf{Venue}} & \multicolumn{2}{c}{\textbf{Unique (\(\sim\)19\%)}} & \multicolumn{2}{c}{\textbf{Multiple (\(\sim\)81\%)}}  & \multicolumn{2}{c}{\textbf{Accuracy}} & \textbf{Inference} \\
~& ~& \textbf{0.25} & \textbf{0.5}& \textbf{0.25} & \textbf{0.5}& \textbf{0.25} & \textbf{0.5} & \textbf{Speed (FPS)}\\
\midrule
\multicolumn{9}{l}{\textbf{\textit{Two-Stage Model} }} \\
\midrule
ScanRefer~\citep{chen2020scanrefer} & ECCV'20 & 76.33& 53.51 &32.73& 21.11& 41.19 & 27.40 & \textbf{6.72} \\
TGNN~\citep{huang2021text} & AAAI'21 & 68.61 &56.80& 29.84 &23.18& 37.37 & 29.70 & 3.19 \\
InstanceRefer~\citep{yuan2021instancerefer} & ICCV'21 & 77.45& 66.83& 31.27 &24.77& 40.23 & 30.15 & 2.33 \\
SAT~\citep{yang2021sat} & ICCV'21 & 73.21 &50.83 &37.64 &25.16 & 44.54 & 30.14 & \underline{4.34} \\
FFL-3DOG~\citep{feng2021free} & ICCV'21 & 78.80& 67.94 &35.19 &25.7 & 41.33 & 34.01 & Not released \\
3D-SPS~\citep{luo20223d} & CVPR'22 & 84.12 &66.72& 40.32 &29.82& 48.82 & 36.98 & 3.17 \\
BUTD-DETR~\citep{jain2022bottom} & ECCV'22 &82.88 &64.98 &44.73 &33.97& 50.42 & 38.60 & 3.33 \\
EDA~\citep{wu2023eda} & CVPR'23 & 85.76 &68.57 &49.13 &37.64 & 54.59 & 42.26 & 3.34 \\
3D-VisTA~\citep{zhu20233d} & ICCV'23 & 77.40& 70.90& 38.70& 34.80& 45.90 & 41.50 & 2.03 \\
VPP-Net~\citep{shi2024aware} & CVPR'24 & 86.05 &67.09& 50.32 &39.03 & 55.65 & 43.29 & Not released \\
\(\text{G}^3\)-LQ~\citep{wang2024g} & CVPR'24 & \textbf{88.09}& \textbf{72.73}& \underline{51.48}& \textbf{40.80}& \underline{56.90} & \textbf{45.58} & Not released \\
MCLN~\citep{qian2025multi} & ECCV'24 & \underline{86.89} &\textbf{72.73} &\textbf{51.96} &\underline{40.76} & \textbf{57.17} & \underline{45.53} & 3.17 \\
\midrule
\multicolumn{9}{l}{\textbf{\textit{Single-stage Model} }} \\
\midrule
3D-SPS~\citep{luo20223d} & CVPR'22 & 81.63 &64.77 &39.48 &29.61 & 47.65 & 36.43 & 5.38 \\
BUTD-DETR~\citep{jain2022bottom} & ECCV'22 & 81.47 &61.24 &44.20 &32.81 & 50.22 & 37.87 & 5.91 \\
EDA~\citep{wu2023eda} & CVPR'23 & 86.40& 69.42& 48.11 &36.82& 53.83 & 41.70 & \underline{5.98} \\
\(\text{G}^3\)-LQ~\citep{wang2024g} & CVPR'24 &\textbf{88.59}& \textbf{73.28}& \underline{50.23} & \underline{39.72} & \underline{55.95} & \underline{44.72} & Not released \\
MCLN~\citep{qian2025multi} & ECCV'24 & 84.43 & 68.36 &49.72 &38.41 & 54.30 & 42.64 & 5.45 \\
TSP3D (Ours) & ----- & \underline{87.25} & \underline{71.41}& \textbf{51.04} &\textbf{42.37}  & \textbf{56.45} & \textbf{46.71} & \textbf{12.43} \\
\bottomrule
\end{tabular}
}
\end{table*}

\section{Qualitative Comparisons}\label{sec:supp4}
\begin{figure*}[t]
	\centering
	\includegraphics[width=0.98\linewidth]{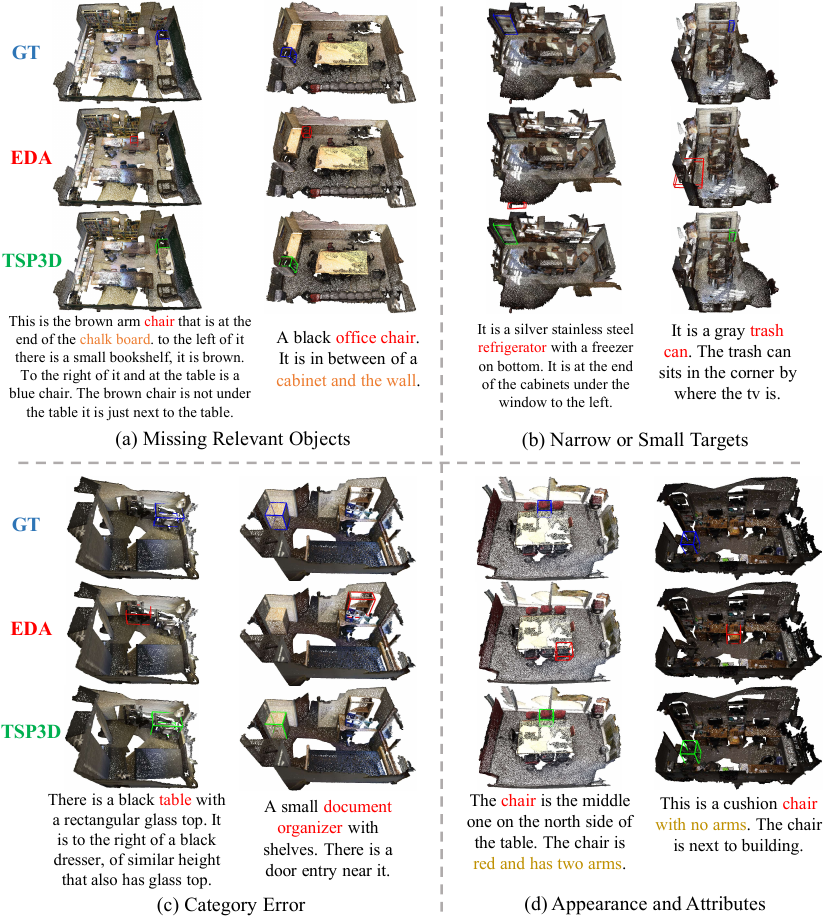}
    \caption{Qualitative results of EDA~\citep{wu2023eda} and our TSP3D on the ScanRefer dataset~\citep{chen2020scanrefer}. In each description, the red annotations indicate the target object. The orange annotations in (a) refer to relevant objects, while the yellow annotations in (d) denote the appearance or attributes of the target. TSP3D demonstrates exceptional performance in locating relevant objects, narrow or small targets, identifying categories, and distinguishing appearance and attributes.}
	\label{fig:supp}
	%\vspace{-.2cm}
\end{figure*}

To qualitatively demonstrate the effectiveness of our proposed TSP3D, we visualize the 3DVG results of TSP3D alongside EDA~\citep{wu2023eda} on the ScanRefer dataset~\citep{chen2020scanrefer}. As shown in Fig.~\ref{fig:supp}, the ground truth boxes are marked in blue, with the predicted boxes for EDA and TSP3D displayed in red and green, respectively. EDA encounters challenges in locating relevant objects, identifying categories, and distinguishing appearance and attributes, as illustrated in Fig.~\ref{fig:supp} (a), (c), and (d). In contrast, our TSP3D gradually focuses attention on the target and relevant objects under textual guidance and enhances resolution through multi-level feature fusion, showcasing commendable grounding capabilities. Furthermore, Fig.~\ref{fig:supp} (b) illustrates that TSP3D performs better with small or narrow targets, as our proposed completion-based addition can adaptively complete the target shape based on high-resolution backbone feature maps.

\section{Limitations and Future Work}\label{sec:supp5}
Despite its leading accuracy and inference speed, TSP3D still has some limitations. First, the speed of TSP3D is slightly slower than that of TSP3D-B. While TSP3D leverages TGP to enable deep interaction between visual and text features in an efficient manner, it inevitably introduces additional computational overhead compared to naive concatenation. In future work, we aim to focus on designing new operations for multi-modal feature interaction to replace the heavy cross-attention mechanism. Second, the current input for 3DVG methods consists of reconstructed point clouds. We plan to extend this to an online setting using streaming RGB-D videos as input, which would support a broader range of practical applications.

% \section{Potential Limitations} 
% Despite of the leading accuracy and inference speed, there are still some limitations of TSP3D. First, the speed of TSP3D is bit slower than TSP3D-B. Although TSP3D utilizes TGP to enable deep interaction between voxel and text features in an efficient way, it unavoidably introduces additional computational overhead compared with naive concatenation. In the future work, we aim to work on designing new operations for multi-modal feature interaction to replace the heavy cross-attention mechanism. Second, currently the input of 3DVG methods is a reconstructed point clouds. We will work on extending it to online setting with streaming RGB-D videos as input, which can support a wider range of practical application.